# Application of artificial neural network to determine the thickness profile of thin film

Archana Bora[1]

*Dept. of Applied Sciences (Physical Science Division), Gauhati University, Guwahati-781014, India*

**Abstract**

In this paper, we introduce a novel artificial neural network (ANN) based scheme to estimate the thickness of thin films deposited on a given substrate. Here we consider the visible interference pattern between a plane wave and a diverging wave reflected from the thin film surface that records the thickness information of the thin film. We assume a uniform thickness profile of the film. However, the thickness increases as the deposition takes place. We extract the intensity data along a line through the center of the interference pattern. We train our network by using a number of such line information of known thickness profiles. The performance of the trained network is then tested by estimating the thickness of unknown surfaces. The numerical simulation results show that the proposed technique can be very much useful for automated measurement of thickness, quickly and in real time, during deposition.

1.  **Introduction**

A thin film is a layer of any material with thickness ranging from fractions of a nanometer to several micrometers. As the thickness of the material decreases compared to the other two dimensions, the surface characteristics dominate the bulk properties of the material and then decide its overall physical and chemical behavior [1]. With the advancement of the thin film technology, now it has become possible to create a wide range of variations in the characteristics of the thin-films by controlling the vital parameters of the growth process paving their way of use in the most technologically advanced applications and industries.

As in such applications, almost all the properties of a particular thin film depend on its thickness, hence an accurate estimate of the thickness has been one of the most important deciding factor in the application of thin films in industrial sectors. Some examples of such sectors are display industry, semiconductor devices, eye glasses, stents, solar cells, polymer coatings, photoresists, solar panels, LCD, MEMS, thin-film packaging etc.

There have been different methods in use for the measurement of the thin film thicknesses. Ion beam analysis, TEM, ellipsometry, surface profilometry etc. are few examples to mention about. In the present work we have proposed to automate the estimation of the thickness of a growing thin-film by applying an artificial neural network (ANN). [2, 3] have measured the thickness and refractive indices for transparent thin films using ANN with spectroscopic reflectometry data. In our work we have demonstrated that an ANN can be used to estimate thickness growth of a thin film based on line profile of interference patterns. The interference is produced between a reference beam and an object beam reflected from the film. While a plane wave is considered as the reference beam, a converging or a diverging beam will serve as an object beam. In the plane of observation, we can record the interference patterns. The shape and the fringe width of the interference pattern thus produced depend on the thickness of the thin-film used, the information about which is carried to the plane of observation by the object beam. Hence the fringes can be used to estimate the thickness of the thin-film specimen.

[1]* Corresponding author. Tel.: +0-361-291-9094 ; fax: +0-000-000-0000 .
*E-mail address:* abora.80@gauhati.ac.in

In the present work we have used such information as an input to a machine learning algorithm to train it. While many different methods are available for the measurement, as has been mentioned above, the main motive of this work is to automatize in-situ thickness measurement process. For this work we have generated interference patterns between a reference beam and an object beam getting reflected from the specimen surface to produce the inputs for the algorithm. For this, we have worked out the necessary expressions to construct the interference pattern numerically and the details of the same are presented in section 2. Based on the formulae we have generated two sets of data, first set for training the ANN and the second set to test the performance of the trained network. Section 3 describes the data generation methods. In section 4 the basic architecture of the ANN has been explained and finally section 5 and 6 present the result and discussions respectively.

## 2. Development of analytical expressions for the interference patterns

Fig 1(i) shows the schematic diagram of a typical experimental set up to generate the interference fringes and Fig1(ii) shows the resultant patterns produced which are concentric circles with decreasing fringe widths. The interference pattern is formed by the superposition of a plane wave (reference beam), and a converging or diverging *object beam* reflected from the top surface of the test thin-film. A beam splitter has been kept in the path of the incident beam at an angle of $45^0$ to the direction of its motion and the plane of observation lie perpendicular to the direction of propagation of the reference beam. Here, the test surface with uniform thickness is in the XY-plan while the observation plane is perpendicular to it.

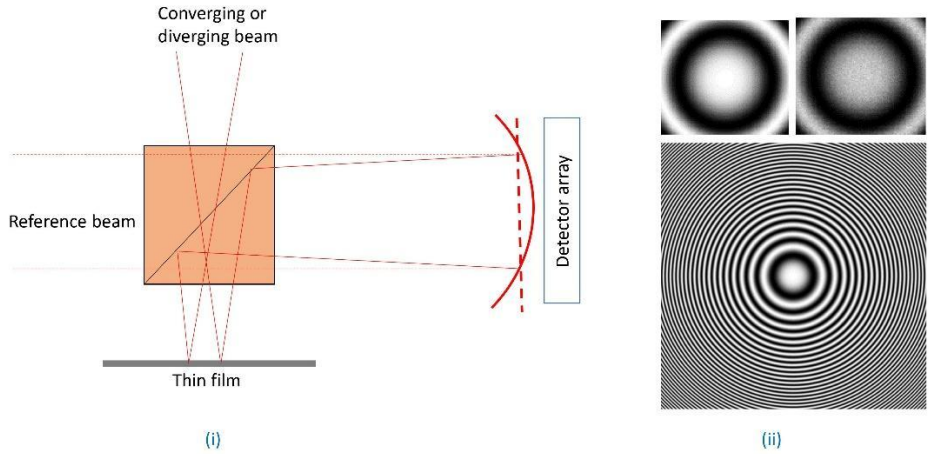

Fig. 1. (i) Schematic diagram of the experimental setup; (ii) Concentric fringes at the detector plane.

Now, let us assume that the incident object beam at the observational plan has a spherical wave front of radius $R_0$ at time *t=0* and its complex amplitude be represented as

$$E_0 = Ce^{i(\omega t - \phi_0)} \qquad \text{(i)}$$

and the corresponding expression for the reference beam is

$$E_r = Ce^{i(\omega t - \phi_r)} \qquad \text{(ii)}$$

As the time dependent components of the phase are the same for both the interfering beams, so only the space dependent parts of phase will have main contribution to the formation of the interference patterns. Thus we will only consider the space dependent phase components only hereafter. The resultant intensity of the interference pattern on the plane of observation will be

$$I = \left| Ce^{-i\phi_0} + Ce^{-i\phi_r} \right|^2$$
$$= C^2 \left| 1 + e^{-i\Delta\phi} \right|^2 \quad \text{(iii)}$$

where, $\Delta\varphi = \frac{2\pi}{\lambda} z_0$ is the resultant phase difference between the two beams [4] . Now, the equation of the spherical surface of the object beam receiving the plane of observation at time $t=0$ can be expressed as

$$x^2 + y^2 + (z - R_0)^2 = R_0^2$$

so that,
$$\Rightarrow z = R_0 - \sqrt{R_0^2 - (x^2 + y^2)} \quad \text{(iv)}$$

where x, y and z correspond to the respective co-ordinates in the observation plane. Therefore, the resultant phase difference can be expressed in terms of the radius of curvature of the spherical wave front as,

$$\Delta\varphi = \frac{2\pi}{\lambda} \left( R_0 - \sqrt{R_0^2 - (x^2 + y^2)} \right) \quad \text{(v)}$$

However, as the thin film deposition starts on the surface of the substrate, the thickness *T* of the film will introduce an additional path difference of *2T* in the object beam. Thus for a growth of *T* thickness of the thin film, the two interfering beams will actually have a phase difference of,

$$\Delta\varphi = \frac{2\pi}{\lambda} \left\{ (R_0 - 2T) - \sqrt{(R_0 - 2T)^2 - (x^2 + y^2)} \right\} \quad \text{(vi)}$$

Now, for a light source the typical wavelength is of order $10^{-7}$cm and a thin film has dimension of $10^{-7}$cm too. In the experimental set up, **R$_o$** has the dimension in the cm range. In such a condition, considering the fixed number of significant figures allowed by a detector system like computer, CCD etc. we can approximate $\frac{(Ro - 2T)^2}{\lambda^2}$ as being equal to $\frac{Ro^2}{\lambda^2}$. Hence, equation (vi) becomes,

$$\Delta\varphi = 2\pi \left\{ \frac{(R_0 - 2T)}{\lambda} - \sqrt{\frac{(R_0 - 2T)^2 - (x^2 + y^2)}{\lambda^2}} \right\}$$

$$= 2\pi \left\{ \frac{R_0}{\lambda} - \frac{2T}{\lambda} - \sqrt{\frac{(R_0 - 2T)^2}{\lambda^2} - \frac{(x^2 + y^2)}{\lambda^2}} \right\}$$

$$= 2\pi \left\{ \frac{R_0}{\lambda} - \frac{2T}{\lambda} - \sqrt{\frac{R_0^2}{\lambda^2} - \frac{(x^2 + y^2)}{\lambda^2}} \right\}$$

$$= 2\pi \left( \frac{R_0}{\lambda} - \sqrt{\frac{R_0^2}{\lambda^2} - \frac{(x^2 + y^2)}{\lambda^2}} \right) - 2\pi \left( \frac{2T}{\lambda} \right)$$

$$= \Delta\phi_{const} - \Delta\phi_{var} \quad \text{(vii)}$$

Where $\Delta\phi_{const} = 2\pi\left(\frac{R_o}{\lambda} - \sqrt{\frac{R_o^2}{\lambda^2} - \frac{x^2+y^2}{\lambda^2}}\right)$ and $\Delta\phi_{var} = 2\pi\left(\frac{2T}{\lambda}\right)$ are the constant part and the thin-film thickness dependent part of the phases. Here, $\Delta\phi_{const}$ is a fixed quantity for a particular experimental set-up and thus can be ignored. The intensity distribution in the observational plane can be obtained by making a substitution for $\Delta\phi$ in equation (iii) as explained above. Interference patterns for different thin films thickness are generated by varying the value of $T$ in the expression of $\Delta\phi_{var}$.

## 3. Generation of train and test set of data

As discussed in the section 2, an interference pattern, formed by superposition between a plane reference wave-front and a spherical wave-front reflected from the top surface of a thin-film preserves the information about the thickness of the film. So, a set of data are generated in Matlab using equation (iii), for different values of $T$ between 5-nm and 200-nm as the deposition takes place. We have recorded the interference pattern at a step size of 10-nm of the thickness. The set is then used to train the ANN. Similarly, another set of data at a step size of 5nm is generated for evaluating the performance of the trained ANN. This set is known as the test set of data.

To generate the data, we express the $T$ in terms of $\lambda$ and consider its value to increase in step of $\lambda/100$. Therefore, if we consider $\lambda=500$ nm then thickness increment ($\Delta T$) is equal to 5nm. The radius of the object beam wave front in the observation plane, $R_o$, is taken as 5 cm. The horizontal co-ordinate of the observation plan (x) is incremented in step of $4\lambda=2\mu m$. For each thickness, interference pattern over 1000 pixels are generated. The intensity data is normalised to vary between 0 and 1.

However, in any electronic sensor like CCD, during information acquisition process random noises get superimposed on the signal. Such noises are best governed by Poisson statistics. Keeping this fact in mind, we have added Poisson noise to the test set of generated data. The maximum grey value ($G_{max}$) possible in an 8-bit detector is 255 while the same for a 10-bit detector is 1023. In order to generate the Poisson noise added data, we have used the built in function *poissrnd* of Matlab. If $I_x$ represents the normalised data without noise, then the corresponding data with Poisson noise is defined as *poissrnd($I_x*G_{max}$)/ $G_{max}$*. Thus Poisson noise added interference data are generated for each thickness values over 1000 pixels.

The signal to noise ratio ($\sigma$) of the Poisson noise added data is $\frac{1}{\sqrt{G_{max}}}$ which is equal to 0.0312 and 0.626 for the 10 bit and 8 grey levels respectively of the detectors. In collecting the data, we consider a line profile of the fringe pattern passing through the center of the pattern. Fig 2 shows such line profiles of the data generated. While Fig 2(i) is the line profile of the fringes without noise, Fig 2(ii) presents the corresponding Poisson noise added data.

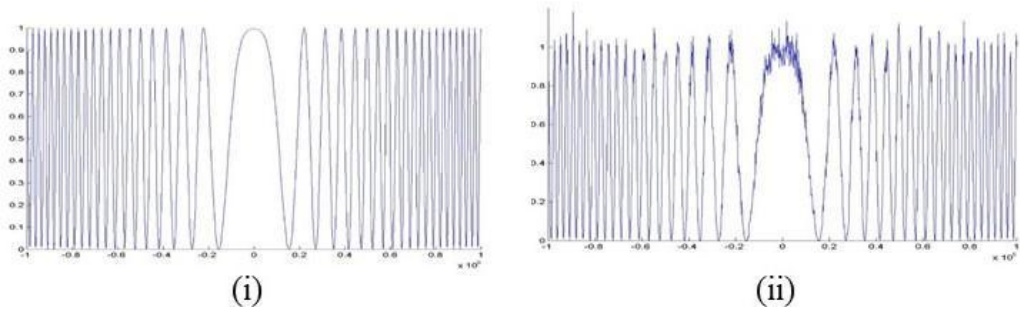

Fig. 2. Line profile of the intensity data of concentric fringes (i) without noise, (ii) with Poisson noise.

## 4. Artificial Neural Network (ANN) Architecture

We have selected artificial neural network as the machine learning algorithm to estimate the thickness growth of a thin film during its formation stage. Supervised training method based on back propagation algorithm [5,6] is used to '*teach*' the network the salient features of the train data set which has been generated as explained in the section 3.

The performance of an ANN mainly depends on the architecture of the network used. In general, an ANN has three different types of layers namely input layer, hidden layers and output layer respectively as shown in Fig 3. The layers are interconnected through different *nodes* and the nodes possess individual random *weight* values at the beginning. The recorded interference patterns to be learned during the training session are feed into the input layer which then get passed on to the hidden layer through the nodes. In the hidden layers the information from the input layer are processed and passed on to the output layer. The output layer includes a set of codes that represent the thickness information of the thin films. In the learning process, the network estimates the differences between this known value at the output layer and the estimated output based on the forward feed from the hidden layer. Then in the next stage, the difference which is known as the *error*, is fed backward through the network and weight associated with each of the nodes gets updated so that the error gets minimize. The whole process is repeated iteratively, until a desired minimum error threshold is reached. In this stage the network finishes learning the data. The weight values, associated with different nodes of the network, get frozen and the network is now ready for the test session. So, when the test patterns are fed to the network, output will be the estimated thickness of the thin-film in terms of the training set. During the learning process, the processing of the input information in the hidden layer takes place through an *activation function*.

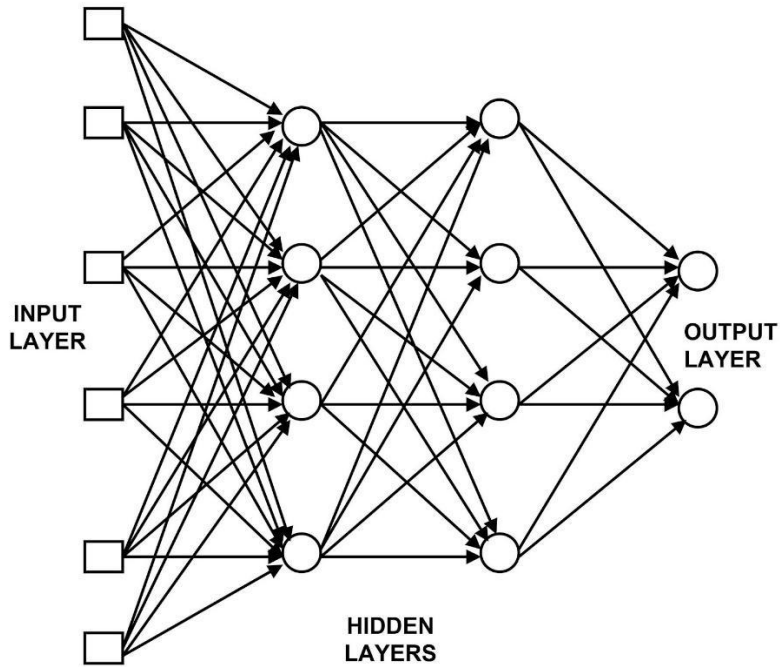

Fig 3: Multilayer-layer feedforward ANN architecture diagram

In our present work, we have used an ANN architecture with two hidden layers with 64 nodes each. The *Sigmoid function* is used as the activation function. In each set of training data with a particular *T* value, there are 40 number of data points representing the line profile of the interference fringe produced by the film. The entire training data set consists of such patterns with 20 different *T* values at step size Δ*T*=10nm. Similarly, the test data set has such 40 different patterns for corresponding *T* values with Δ*T* = 5nm.

## 5. Results

The test data set as described in section 3 is employed in our ANN to estimate the thickness of the film and the results are presented as a scatter plot in Fig4. In the plots, the actual thickness of the films is denoted as "Catalogue" while the ANN estimated thickness is denoted as "ANN". The test data set comprises both the Poisson noise added interference patterns correspond to the 8-bit and 10-bit detector systems.

The scatter plots of the classification results corresponding to 8-bit grey level is presented in Fig.4(i) while the same for the 10-bit grey level is shown in Fig 4(ii) respectively. Both the X and Y-axis labels are expressed in nm unit. The signal to noise ratio ($\sigma_{noise}$) corresponding to the test data set are also shown in the respective figures.

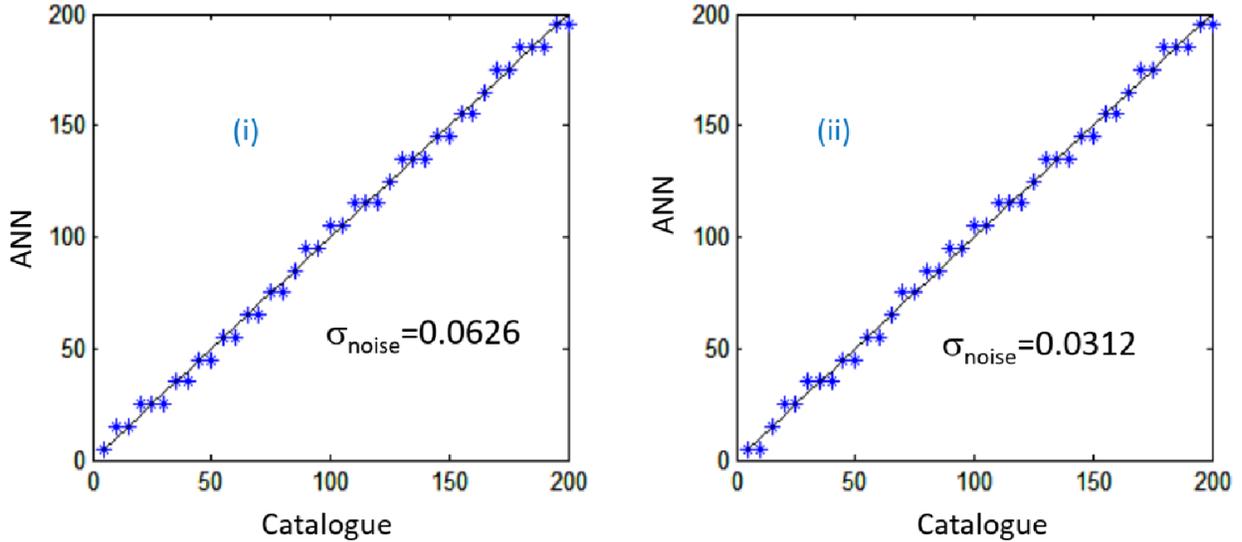

Fig 4: The scatter plots of the classification results with (i) 8- bit detector and (ii) 10- bit detector. The horizontal and the vertical axes are expressed in nm unit.

## 6. Discussion

The results presented in the above section clearly demonstrates that the proposed ANN based scheme is able to estimate the thickness of the film from the 1-D interference data with reasonable accuracy. The classification accuracy in terms of the root mean square (RMS) error for both the detectors is found to be 0 .7 nm. It can be seen from the figures that the ANN has detected the thickness correctly in most of the cases except for few outliers. Careful observation of the outliers show that they belong to intermediate thickness values of the training session data. However, the most important finding of the work is that though the network was trained using pure line profile of interference fringes, however it can efficiently classify the Poisson noise added data as well. It is evident that the classification process of the ANN is not affected by the change in the Poisson noises as indicated by the constant RMS error in the scatter plots. Considering the affordability and commercial availability, we have considered detector noise corresponding to 8 bit and 10-bit array detectors. However, the scheme is expected to be applicable to higher grey level detectors as well. We have considered illumination wavelength of 500 nm. Nevertheless, the same can be replaced by a lower wavelength using a laser emitting violet or blue light, in which case the accuracy of the estimation will be improved even further. It is to be noted that the proposed ANN based scheme is suitable for real time in-situ measurement of film thickness where the interference data is available since the beginning of the deposition process. In order to be useful for ex-situ measurement the scheme will require the interface data corresponding to the bare substrate.